\documentclass[11pt]{article}

\usepackage[]{acl}

\usepackage{times}
\usepackage{latexsym}

\usepackage[T1]{fontenc}
\usepackage[utf8]{inputenc}

\usepackage[most]{tcolorbox}
\usepackage{arydshln}
\usepackage{array}
\usepackage{xspace}

\newcolumntype{L}[1]{>{\raggedright\let\newline\\\arraybackslash\hspace{0pt}}m{#1}}
\newcolumntype{C}[1]{>{\centering\let\newline\\\arraybackslash\hspace{0pt}}m{#1}}
\newcolumntype{R}[1]{>{\raggedleft\let\newline\\\arraybackslash\hspace{0pt}}m{#1}}

\makeatletter
\def\adl@drawiv#1#2#3{%
        \hskip.5\tabcolsep
        \xleaders#3{#2.5\@tempdimb #1{1}#2.5\@tempdimb}%
                #2\z@ plus1fil minus1fil\relax
        \hskip.5\tabcolsep}
\newcommand{\cdashlinelr}[1]{%
  \noalign{\vskip\aboverulesep
           \global\let\@dashdrawstore\adl@draw
           \global\let\adl@draw\adl@drawiv}
  \cdashline{#1}
  \noalign{\global\let\adl@draw\@dashdrawstore
           \vskip\belowrulesep}}
\makeatother

\makeatletter
\DeclareRobustCommand\onedot{\futurelet\@let@token\@onedot}
\def\@onedot{\ifx\@let@token.\else.\null\fi\xspace}

\makeatother

\newcommand{\ours}{\textsc{SenseShift}\xspace}

\newcommand{\mb}{\textsc{ModernBert}\xspace}
\newcommand{\mbs}{\textsc{\mb-D-28m}\xspace}
\newcommand{\mbm}{\textsc{\mb-E-0.15b}\xspace}
\newcommand{\mbl}{\textsc{\mb-E-0.4b}\xspace}
\newcommand{\gemmas}{\textsc{Gemma2-2b}\xspace}
\newcommand{\gemmam}{\textsc{Gemma2-9b}\xspace}
\newcommand{\lms}{\textsc{Llamma3.2-3b}\xspace}
\newcommand{\lmm}{\textsc{Llamma3.1-8b}\xspace}
\newcommand{\gpts}{\textsc{GPT4o-mini}\xspace}
\newcommand{\ossm}{\textsc{OSS-20b}\xspace}
\newcommand{\ossl}{\textsc{OSS-120b}\xspace}
\newcommand{\qwm}{\textsc{Qwen3-32b}\xspace}
\newcommand{\gptmini}{\textsc{GPT4o-Mini}\xspace}

\usepackage{parskip}%
\usepackage{graphicx}%
\usepackage{siunitx}
\usepackage{booktabs}
\usepackage{float}
\usepackage{subcaption}
\usepackage[table]{xcolor}
\usepackage{colortbl}
\definecolor{g}{RGB}{170, 220, 170}
\definecolor{r}{RGB}{250, 170, 170}
\usepackage{microtype}

\usepackage{inconsolata}

\usepackage{graphicx}

\usepackage{multirow}

\title{\ours: Continuous Sentiment-Controlled Text Generation via Encoder-based Mask Infilling}

\author{
    \textbf{Shahed Masoudian\textsuperscript{1}}, 
    \textbf{Markus Frohmann\textsuperscript{2,3}}, 
    \textbf{Emmanouil Karystinaios\textsuperscript{1}},\\ 
    \textbf{Navid Rekabsaz\textsuperscript{2}}, 
    \textbf{Markus Schedl\textsuperscript{1}} \\
  \textsuperscript{1} Johannes Kepler University Linz \\
  \textsuperscript{2} Thomson Reuters Labs \\
  \textsuperscript{3} University of Toronto, Vector Institute \\
  \small{
    \textbf{Correspondence:} \href{mailto:}{shahed.masoudian@jku.at}
  }
}

\begin{document}
\maketitle
\begin{abstract}

Recent controllable text generation (CTG) for sentiment control has largely focused on decoder-based large language models, making causal attention the dominant paradigm. While effective for fluent generation, these models still struggle to satisfy complex constraints and follow fine-grained sentiment signals specified by users. Existing sentiment-aware CTG methods typically simplify the problem by treating sentiment either as a coarse categorical label (e.g., positive or negative) or as a single fine-grained control signal applied to an entire document. Consequently, more challenging settings such as sentence-level sentiment control within long-form text remain underexplored. To address these limitations, we introduce \ours, an encoder-based framework for fine-grained sentence-level CTG. Unlike standard decoder architectures, \ours leverages bidirectional attention, quantized sentiment signals, and iterative mask infilling to generate local sentences conditioned on target sentiment intensity. Empirical evaluations on story and review generation demonstrate that \ours achieves stronger sentiment controllability while maintaining text quality and robustness to out-of-domain generation compared to larger decoder-based baselines.~\footnote{The source code for experimental setup and training is available at \url{https://github.com/ShawMask/SenseShift}, and the model checkpoints for inference can be found at \url{https://huggingface.co/shawhed/SenseShift-large}.}

\end{abstract}

\section{Introduction}
\label{sec:introduction}

Controllable Text Generation (CTG) aims to steer generated text toward desired attributes specified using external conditions~\cite{zhang-etal-2023-survey}. In sentiment-aware CTG, the goal is to regulate the emotional tone of text based on users desire~\cite{lorandi-belz-2023-control}. Prior works investigated categorical sentiment control for desired segments~\cite{Yuan-etal-2022-hierarchical, liang_controlled_2024} and fine-grained sentiment control of full-document~\cite{luo_towards_2019, tang_latent_2021}. These works predominately rely on decoder LLMs~\cite{yang_plug-language_2024}, making causal attention a default paradigm for sentiment-aware CTG. We argue that this choice of architecture is not best fit for scenarios where users wish to rewrite sentences of a document with a desired fine-grained sentiment value, an area still underexplored.

Recent studies show that decoder-only architectures are fundamentally constrained by their unidirectional information flow~\cite{kopiczko-etal-2025-bitune}. Moreover, CTG via prompting requires precise task definition which often result in unstable or weakly controlled outputs, especially under ambiguous or subtle sentiment targets~\cite{labroo-etal-2026-funny, laban_llms_2025, pmlr-v139-zhao21c}. In particular, controlling sentiment \emph{intensity} even as categorical attribute (e.g., \textit{happy} vs \textit{very happy}) remains challenging for prompting-based approaches. We provide empirical evidence for this limitation in our task setting in Appendix~\ref{sec:app:gpt_story_generation}.

\begin{figure*}[t]
    \centering
    \includegraphics[width=1\textwidth]{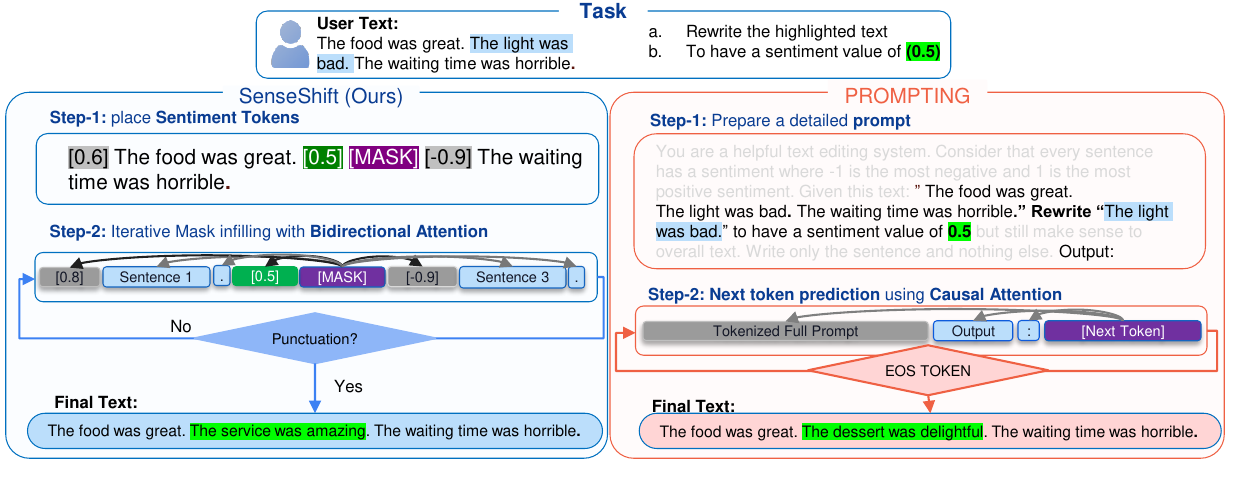}
    \caption{\ours VS decoder-based prompting for sentiment CTG. Based on a given text, target sentence and sentiment, \ours places sentiment tokens before each sentence. encoder-architecture allows \ours to predict masked tokens attending to each target sentiment while distinguishing prior context from posterior.}
    \label{fig:flowchart}
\end{figure*}

To address these limitations, we introduce \ours, a novel encoder framework for fine-grained sentiment-aware in-text generation. Unlike decoder-based approaches that rely on causal attention and generate strictly left-to-right, \ours leverages bidirectional attention together with iterative mask infilling. This enables in-text generation conditioned jointly on prior and posterior context, so that generated sentences integrate more naturally with the surrounding text. \ours consists of three stages: automatic sentiment signal construction and quantization, control-aware MLM fine-tuning, and iterative mask infilling at inference. Together, these enable fine-grained manipulation of sentiment in arbitrary sentences within a long text. Figure~\ref{fig:flowchart} illustrates how \ours generation differs from decoder prompting.\footnote{Qualitative examples of text modified with \ours can be found in Figures~\ref{app:fig:story_qualitative} and~\ref{app:fig:review_qualitative}.}

% To address these limitations, we introduce \ours, a novel encoder framework for continuous sentiment controllable text generation. Unlike decoder-based approaches that rely on causal attention and generate strictly left-to-right, \ours leverages bidirectional attention together with iterative mask infilling. This enables in-text generation conditioned jointly on prior and posterior context, allowing sentences to integrate naturally with the surrounding text. \ours include three stages: automatic sentiment signal construction and quantization, control-aware MLM fine-tuning, and iterative mask infilling at inference, which together enables fine-grained manipulation of arbitrary sentences of a long text. Figure~\ref{fig:flowchart} illustrates how \ours generation differs from decoder prompting.\footnote{Qualitative examples of text modified with \ours can be found in Figures~\ref{app:fig:story_qualitative} and~\ref{app:fig:review_qualitative}.}

We instantiate \ours on \mb~\cite{warner2024modernbert} \textit{base} and \textit{large} substantially smaller than the decoder baselines. Through training and evaluation on review and story-writing datasets we demonstrate empirically that \ours consistently outperforms decoder-based baselines including prompting, activation steering, and instruction tuned models many times its size on sentiment control while maintaining fluency and contextual fit. Human evaluation confirms this: annotators judge \ours outputs slightly more fit to context compared to a larger decoder baseline model while also preferring \ours for sentiment alignment. Together, these findings suggest that encoder-based generation is an effective and competitive alternative to autoregressive decoding for fine-grained sentiment-aware in-text generation.

\section{Related Work}
\label{sec:related_work}

Early approaches to sentiment and style control used latent-variable or recurrent formulations to inject attribute information during generation~\cite{hu-etal-2017-toward, peng_towards_2018}. These works assume that sentiment is a categorical label and steer generation toward those classes, for tasks such as poetry generation~\cite{shao_sentiment_2021}, multi-aspect control over sentiment and topic~\cite{ding-etal-2023-maclasa}, reinforcement-style unlearning~\cite{lu_quark_2022}, and contrastive prefix methods~\cite{qian_controllable_2022, zheng-etal-2023-click}. Decoding-time steering methods~\cite{pascual-etal-2021-plug-play, krause-etal-2021-gedi-generative, liu-etal-2021-dexperts, yang-klein-2021-fudge} and supervised fine-tuning approaches~\cite{yang_plug-language_2024, zal_sentiment_2024} share the same coarse-grained, document-level limitation. Across all these lines, sentiment is treated as a polarity signal and control operates over full sequences.

A smaller body of work has explored fine-grained control signals, including early work on controlling gender information within encoders for fairness~\cite{masoudian-etal-2024-effective} using gated adapters.  Subsequent works in decoder-based generation focuse on fine-grained sentiment transfer via Gaussian kernel conditioning on standalone sentences~\cite{luo_towards_2019}, scorer-based unsupervised control~\cite{jain_unsupervised_2019}, emotional control in story generation via decoder-side guidance~\cite{wang_chae_2022}, and continuous interpolation mechanisms~\cite{kangaslahti_continuous_2024, samuel_cie_2025}. These methods target sentiment granularity but operate at the document level (i.e., ignoring sentiment dynamics).

The works structurally most similar to ours introduce positional awareness into CTG. Story-ending control uses fine-grained sentiment and Gaussian kernel conditioning to guide only the concluding segment of stories~\cite{luo-etal-2019-learning}. Other methods use conditional variational auto-encoder, aspect-level disentanglement, hierarchical templates, and dynamic attribute graphs to control sentiment of parts of a text~\cite{qiao_sentiment-controllable_2020, Yuan-etal-2022-hierarchical, nawezi_style_2023, liang_controlled_2024}. These works demonstrate the value of structural awareness but use categorical sentiment and do not move beyond decoder-based architecture.

From an architectural perspective, \ours is most closely related to the text infilling paradigm~\cite{donahue-etal-2020-enabling}. Unlike their work, which fine-tunes a decoder model on masked text without attribute signal, \ours leverages the encoder that attends jointly to both surrounding context and a quantized sentiment intensity signal before generating the target sentence.

\section{Methodology}
\label{sec:method}

We aim to have sentence-level and fine-grained control over the sentiment of generated text which is particularly suitable for encoders. Their bidirectional attention and masked token prediction allow them to generate within a long context, a special feature that autoregressive decoders do not inherently have~\cite{kopiczko-etal-2025-bitune}. \ours achieves this goal using the following steps:

\subsection{Automatic Sentiment Signal Construction and Quantization}
\label{subsec:signal}

First key obstacle for fine-grained sentiment control is the lack of high-quality labeled sentence-level sentiment in long-form corpora, which can be used to train the models. We automatically derive sentiment scores using VADER~\cite{vader2016} as done in prior works~\cite{konen_style_2024} with one major difference: We extract the sentiment scores per sentence. We chose VADER over more accurate but slower alternatives such as RoBERTa~\cite{barbieri-etal-2020-tweeteval} because it is fast, deterministic, and derived from human annotation, making it practical for large-scale pre-processing and online inference without adding additional complexity, parameters, and latency.

Given a document $\mathcal{D}$ segmented into $m$ sentences $\mathcal{D} = \{s_1, s_2, \dots, s_m\}$, we compute for each sentence $s_i$ the sentiment score $v_i \in [-1,1]$. Scores are then quantized to a discrete control grid with step size 
$$\delta=0.1: \sigma_{s_i} = \operatorname{round}\!\left(\frac{v_i}{\delta}\right)\cdot \delta$$
Each quantized value is mapped to a unique sentiment token $[\sigma_{s_i}]$ and placed at the beginning of its respective sentence. This converts unlabeled document into sentence-level controllable training data, providing the model with local sentiment intensity signals rather than a single document-level sentiment.

\subsection{Control-Aware MLM Fine-Tuning}
\label{subsec:training}

Training \ours is standard Mask Language Modeling (MLM) under two control-aware modifications. First, sentiment tokens $[\sigma_{s_i}]$ are never masked and remain visible throughout training as persistent conditioning anchors.\footnote{Initial experiments revealed that when sentiment tokens are trainable, overall performance degrades, as the model tends to reproduce them rather than using them as generation constraints.} Second, we increase the masking ratio to $40\%$, to encourage stronger dependence on both surrounding context and sentiment anchors when reconstructing masked content rather than relying on local co-occurrence patterns.

\subsection{Iterative Mask Infilling}
\label{subsec:iter_mask_filling}

Predicting all masked tokens in parallel often results in repetitive and incoherent generations, since top token probabilities are predicted independently without sequence-level awareness (see Appendix~\ref{sec:app:iter_v_noniter}). To address this, we mimic iterative generation and beam search generation of decoders and implement a simplified iterative mask infilling approach~\cite{google2016, diversebeam2016} inspired by prior works.
% treating regeneration as iterative masked infilling~\cite{google2016, diversebeam2016}, where tokens are predicted sequentially, each conditioned on previously predicted tokens and the full surrounding document context.

During generation, the original sentiment token of a target sentence $[\sigma_{s_i}]$ is replaced with a user-defined target sentiment $[\sigma'_{s_i}]$. The target sentence $s_i$ is fully corrupted with $[\text{MASK}]$ token, producing a masked input $X'_{masked}$. Starting from the first masked position after the sentiment token, \ours predicts one token at a time. At each step~$t$, the model computes a probability distribution over the vocabulary given the current context $Z_t$, which includes the surrounding document, previously predicted tokens, and the target sentiment signal: $$P(\cdot \mid Z_t) = \text{Softmax} \left( \frac{f_{\theta}(Z_t)}{\tau} \right)$$
where $f_{\theta}(Z_t)$ is the model's logit vector and $\tau$ is a temperature parameter controlling output sharpness.

\paragraph{Beam Search and Scoring.}
To explore the high-probability candidate space, we maintain $B$ active beams from the top-$k$ tokens. Candidates are evaluated using a length-normalized log-probability score to prevent bias toward shorter completions: 
$$S = \frac{1}{\text{LP}(n)} \sum_{t=1}^{n} \log P(w_t \mid Z_t) $$
where $\text{LP}(n) = \frac{(5 + n)^{\alpha}}{(5 + 1)^{\alpha}}$.
Here $n$ is the number of predicted tokens, $\alpha$ is the length penalty hyperparameter, and 5 is a smoothing constant following~\citet{google2016}. To prevent beams from collapsing into repetitive outputs, we apply a diversity penalty $\gamma$ that discourages multiple beams from selecting the same token. Formally, when scoring token $w$ for beam $b$, we count how many higher-ranked beams $b' < b$ have already selected $w$ at the current step and penalize accordingly: 
$$S_{\text{final}} = S - \gamma \cdot \sum_{b' < b} \mathbf{1}[w_{b'} = w]$$.

A beam is marked complete when the predicted token belongs to the terminal punctuation set,\footnote{Punctuation tokes are \{. , ; , ! , ?\}.} at which point any remaining $[\text{MASK}]$ tokens in that beam are replaced with $[\text{PAD}]$ tokens. 
Generation is also capped at a maximum of 30 tokens to prevent degenerate outputs.\footnote{This threshold can alternatively be set to the original sentence length to enforce length-matched generation.} The resulting sentence $s'_i$ is conditioned on the surrounding bidirectional context $\mathcal{D} \setminus \{s_i\}$, the neighboring sentiment signals, and the target constraint $\sigma'_{s_i}$.

\section{Experimental Setup}
\label{sec:ex_setup}

We deployed \ours on \mb~\cite{warner2024modernbert} in two sizes namely \textit{\mbm} (base, 149M parameters and \textit{\mbl} (large, 395M parameters), a pre-trained bidirectional encoder we preferred over predecessors such as BERT~\cite{devlin-etal-2019-bert} and RoBERTa~\cite{liu2019roberta} due to its significantly higher contextual length and modernized training recipe. Regardless, \ours framework is considered independent of model and can be deployed on any other encoder model with Masked Language Modeling. For evaluation we compared \ours across three baseline categories: prompting, activation steering, and fine-tuning methods, all of which we introduce in detail in the following. To assess our method, we train and evaluate on two distinct datasets, measuring both in-domain performance and out-of-domain (OOD) performance. The latter specifically checks for robustness and is relevant for prompting, which by nature operates domain-agnostic. Our dataset selection ensures that both text structure and sentiment dynamics vary throughout the text.

\textbf{TinyStories}~\cite{eldan2023tinystories} is a corpus of $\sim$4.5M short stories generated by GPT-3.5/4.0. The stories are two to three paragraph while including various arcs (e.g., bad ending, inclusion of a twist, dialogues) resulting in sentiment changes.

\textbf{Yelp Reviews}~\cite{zhangCharacterlevelConvolutionalNetworks2015} $\sim$300K samples of human-written reviews spanning multiple service domains with user ratings. We discard user ratings and capture sentence-level sentiments (e.g., praising food quality while criticizing service) using the approach described in Section~\ref{subsec:signal}.

We remove formatting artifacts, discard single-sentence documents, and exclude texts exceeding 500 words, allowing high-quality, structurally complete data. From each dataset, 2000 samples are held out as test sets, unseen during training/validation. \ours is trained/validated independently on each dataset but tested on both. Training hyperparameters, sentiment distribution and datasets statistics, are reported in Appendix~\ref{sec:app:training}. 

\subsection{Baselines}

We compare \ours against four controllable generation paradigms, covering the main families of LLM-based sentiment control. Baselines include open-weight models: \mbs (28M)\footnote{A small version of \mb architecture trained from scratch for next token prediction with causal attention.}, LLaMA Instruct (3.2-3B, 3.1-8B), Gemma2 Instruct (2B, 9B), Qwen 3 Reasoning (32B), and OSS model (20B, 120B), as well as a closed-source \gptmini. Due to GPU memory constraints models up to 3 billion parameters were trained for all controlling methods while larger models are only controlled using zero-shot prompting. Details of prompts and instructions can be found in Section~\ref{subsec:app:prompts}. The sentiment controlling methods are as follow:

\textbf{Prompting}. We asked models to replace a randomly selected sentence with one sentence matching a numerically specified target sentiment.

\textbf{Instruction Tuning}. Fine-tunes decoder models specifically for sentence generation given a target sentiment score. Models are trained to replace a \texttt{[Missing-Sentence]} with a substitute.

\textbf{Token Instruction Tuning}. Extends instruction tuning with the quantized sentiment tokens from \ours (Section~\ref{sec:method}), conditioning generation on the target sentiment at inference. This decoder baseline shares \ours's sentiment representation, isolating the effect of bidirectional infilling.

\textbf{Activation Steering}. Identifies and steers the activation vectors of models without training, following~\citet{rimsky-etal-2024-steering}. During decoding, internal activations are steered along a sentiment axis learned from contrastive sentence pairs, to control generated output without fine-tuning. Full implementation details are provided in Appendix~\ref{subsec:app:steering}.

\subsection{Evaluation Metrics}
\label{subsec:evalation_metric}
We utilize standardized evaluation protocol applied identically to all baselines and \ours. For each document, we randomly select a sentence and assign it a random target sentiment. Models then generate the sentence conditioned on the target. All generated outputs are scored using VADER, the same sentiment analyzer used for training of all baselines, ensuring a consistent and unbiased comparison across baselines (except for zero-shot prompting). Our metrics capture two key aspects: (1) adherence to the target sentiment signal, and (2) fluency and fitness of the generated sentence with its surrounding context.

\textbf{Delta Sentiment ($\Delta_s$)}. Measures the mean absolute error between the target sentiment and the generated sentence sentiment. For a sentence $s_i$ from document $j$, this is defined as
$$\Delta_{s_i} = |\hat{\sigma}_{s_i} - \sigma'_{s_i}|$$
where $\hat{\sigma}_{s_i}$ is the new sentiment of the generated sentence and $\sigma'_{s_i}$ is the target sentiment.

\textbf{Sentiment Accuracy ($Acc.$)}. We evaluate sentiment accuracy based on standard positive, neutral, and negative categories. We define neutral band as $[-0.2, 0.2]$ (wider compared to VADER threshold of $\pm 0.05$~\cite{vader2016}) to avoid considering near-neutral intensity targets as failed control, and uniformly across all baselines and original sentences. Scores higher and lower than this range are labeled  positive or negative respectively.

\textbf{Correlation ($Corr.$)}. measures the Pearson correlation between target sentiment $\sigma'_{s_i}$ and generated sentence sentiment $\hat{\sigma}_{s_i}$. A value of 1.0 indicates perfect linear adherence to the control signal, while 0.0 indicates complete failure in sentiment steering.

\textbf{Perplexity ($PPL.$)}. is computed at the sentence level using GPT-2 and serves as a standard proxy for fluency and coherence. It measures the predictability of generated tokens but does not capture alignment with the surrounding document context, motivating the following metric.

\textbf{Fitness ($\Delta_f$)}. Standard n-gram metrics such as BLEU or ROUGE are not the best choice to measure contextual fitness because the generated sentence may intentionally diverge from the original sentence through sentiment manipulation. Hence, we propose a composite metric combining semantic similarity and entity-level overlap. For a generated sentence $s'_i$ at position $i$, we compute cosine similarity scores using Sentence-BERT embeddings~\cite{reimers-gurevych-2019-sentence}\footnote{Specifically \textit{all-mpnet-base-v2}.} for the preceding context $s_{<i}$ and following context $s_{>i}$ separately. Boundary sentences only have single similarity: $CS_{<i}$ alone for the final sentence and $CS_{>i}$ alone for the first sentence. 

We complement the similarity score with an \textit{entity overlap} score that checks whether named entities in the generated sentence are grounded in the surrounding document. Formally we write:
$$EC_i = \frac{|\text{Entities}(s'_i) \cap \text{Entities}(\mathcal{D} \setminus \{s_i\})|}{|\text{Entities}(s'_i)|}$$
and when $s'_i$ contains no named entities, we set $EC_i = 1$, since the absence of entities introduces no grounding violation. The fitness score for a non-boundary sentence combines the two similarity scores with the entity overlap score, weighting the latter by 2 so that entity grounding contributes on par with the aggregate similarity signal: 
$$C_i = \frac{CS_{<i} + CS_{>i} + 2 \cdot EC_i}{4}$$

Entity overlap flags fabricated entities but does not verify that entities are used in contextually appropriate relations, which we leave to the similarity terms.

For every document we compute the fitness score for the generated and original sentence, and report the calibrated fitness score as: $$\Delta_{f_i} = C(s'_i) - C(s_i)$$
A value of zero indicates no relative change in contextual fit, while a positive value indicates improved fitness, and a negative value indicates degradation. We assess whether $\Delta_{f_i}$ significantly diverges from zero using a two-sided t-test, with results reported in Section~\ref{sec:results}.

\textbf{Human Evaluation}. We complement automatic evaluation with a user preference study comparing \ours against the closest competitive baseline, namely \gemmas trained with token instruction tuning. Eight annotators evaluated 100 parallel documents (55 stories and 45 reviews) generated by both models using the same target sentence and sentiment intensity. They selected their preferred anonymized output based on three quality criteria: \textit{contextual fit} (fitness with the surrounding text), \textit{grammatical fluency} (sentence fluency independent of context), and \textit{sentiment following} (alignment with the target sentiment value on a $-1$ to $+1$ scale). For each criterion, annotators chose among four options namely \textit{A}, \textit{B}, \textit{Both}, or \textit{Neither} without knowing which model produced each sentence. We report the preference percentages in Section~\ref{sec:results}, with full details provided in Appendix~\ref{bsec:app:human_eval}.

\begin{table*}[!ht]
\centering
\small
\setlength{\tabcolsep}{5pt}
\caption{Overall comparison between baselines and \ours. Significantly different changes ($p$-value<0.01) compared to the 0 baseline are marked with an asterisk. Note that columns under green are same models trained on story and columns under red are same models trained on review. in table $PPL_{old}$ refers to perplexity of the original sentence calculated using GPT-2 and $PPL_{new}$ refers to generated sentence perplexity}
\label{tab:control_results_id_ood}
\resizebox{1.0\textwidth}{!}{%
\begin{tabular}{l|cccccc|cccccc}
\toprule
  \multirow{2}{*}{\textbf{Model}} & \multicolumn{6}{c|}{\textbf{Story Evaluation}} & \multicolumn{6}{c}{\textbf{Review Evaluation}} \\
  
 &\textbf{PPL$_{old}$}& \textbf{PPL$_{new}\downarrow$} &\textbf{$\Delta_s$ } $\downarrow$ & $\Delta_f$ $\uparrow$ & \textbf{Corr.} $\uparrow$ & \textbf{Acc.} $\uparrow$ &\textbf{PPL$_{old}$}&\textbf{PPL$_{new}\downarrow$} & \textbf{$\Delta_s$} $\downarrow$ & $\Delta_f$ $\uparrow$ & \textbf{Corr.} $\uparrow$ &\textbf {Acc.} $\uparrow$ \\
\midrule
\textbf{Prompting}&&&&&&&&&&\\
\mbs  &80.6& 255.5 & 0.56 & -0.29* & 0.07 & 0.36 &222.3& 161.2 & 0.69 & -0.13* & -0.19 & 0.23 \\
\lms  &81.1& 118.1 & 0.50 & -0.05* & 0.25 & 0.41 &239.4& 113.1 & 0.51 & -0.00 & 0.21 & 0.38 \\
\gemmas  &88.6& 139.2 & 0.56 & -0.00 & 0.15 & 0.37 &245.1& 117.3 & 0.55 & 0.02* & 0.14 & 0.37 \\
\gemmam  &83.6& 126.2 & 0.39 & -0.03* & 0.65 & 0.55 &268.3& 97.8 & 0.44 & 0.00 & 0.53 & 0.50 \\
\lmm  &73.4& 92.7 & 0.48 & -0.05* & 0.34 & 0.44 &236.2& 81.7 & 0.49 & -0.00 & 0.27 & 0.41 \\
\gpts  &82.6& 82.6 & 0.44 & -0.00 & 0.53 & 0.54 &253.8& 159.7 & 0.42 & -0.00 & 0.48 & 0.49 \\
\ossm  &74.2& 97.5 & 0.31 & -0.04* & 0.75 & 0.64 &236.2& 107.1 & 0.34 & -0.00 & 0.68 & 0.59 \\
\qwm   &73.1& 81.6 & 0.32 & -0.08* & 0.76 & 0.65 &229.3& 92.1 & 0.35 & -0.01 & 0.66 & 0.58 \\
\ossl   &72.1& 110.2 & 0.30 & -0.03* & \textbf{0.793} & 0.66 &236.0& 143.4 & 0.31 & -0.00 & \underline{0.74} & 0.64 \\
\midrule
\midrule
\multicolumn{1}{l}{\textbf{In-Domain Evaluation}}&\multicolumn{6}{c}{\cellcolor{g}\textbf{Story Trained}}&\multicolumn{6}{c}{\cellcolor{r}\textbf{Review Trained}} \\
\midrule
\textbf{Instruction Tune}&&&&&&&&&&&&\\
% \cellcolor{g}
\mbs  &82.2& 402.7 & .43 & -0.04* & 0.43 & .49 &285.7& 140.7 & 0.35 & -0.02 & 0.62 & 0.59 \\
\lms  &80.7& 75.4 & 0.50 & -0.03* & 0.23 & 0.44 &219.5& 81.6 & 0.45 & 0.04* & 0.41 & 0.50 \\
\gemmas  &77.9& 75.5 & 0.43 & -0.03* & 0.44 & 0.49 &276.1& 88.5 & 0.43 & -0.01 & 0.45 & 0.53 \\
\midrule
\textbf{Token Instruction Tune}&&&&&&&&&&&&\\
\mbs  &79.1& 318.0 & 0.35 & -0.17* & 0.57 & 0.54 &266.3& 207.8 & 0.43 & -0.07* & 0.42 & 0.41 \\
\lms  &80.3& 77.6 & 0.28 & \textbf{-0.00} & 0.71 & 0.66 &217.8& 87.2 & 0.32 & 0.04* & 0.64 & 0.62 \\
\gemmas  &77.9& 72.1 & \underline{0.21} & -0.03* & 0.74 & \underline{0.75} &285.5& 108.4 & 0.30 & -0.01 & 0.64 & 0.67 \\
\midrule
\textbf{Steering}&&&&&&&&&&&&\\
\mbs &82.6& 108.2 & 0.55 & -0.30* & 0.04 & 0.31 &241.8& 186.2 & 0.54 & -0.00* & 0.095 & 0.34 \\
\lms &78.3&  221.3 & 0.54 & -0.10* & 0.08 & 0.32 &255.1&186.2 &  0.54 & -0.00* & 0.095 &0.34 \\
\gemmas &85.3& 135.0 & 0.56 & -0.01 & 0.18 & 0.39 &229.6& 124.0 & 0.54 & 0.03* & 0.17 & 0.36 \\
\midrule
\textbf{\ours}&&&&&&&&&&&&\\
\mbm   &78.3& 91.6 & 0.26 & -0.00 & 0.77 & 0.69 &260.3& \underline{55.3} & 0.29 & 0.06* & 0.71 & 0.66 \\
\mbl &81.7& \textbf{53.6} & \textbf{0.20} & \underline{-0.00} & \underline{0.790} & \textbf{0.78} &261.5& 57.5 & 0.26 & 0.04* & 0.71 & 0.69 \\
\midrule
\midrule
\multicolumn{1}{l}{\textbf{Out of Domain Evaluation}}&\multicolumn{6}{c}{\cellcolor{r}\textbf{Review Trained}}&\multicolumn{6}{c}{\cellcolor{g}\textbf{Story Trained}} \\
\midrule

\textbf{Instruction Tune}&&&&&&&&&&&&\\
\mbs  &81.0& 366.0 & 0.35 & -0.24* & 0.61 & 0.57 &251.5& 614.1 &0.49 &-0.03* & 0.30 & 0.42 \\
\lms &81.2& 86.6 & 0.48 & -0.04* & 0.35 & 0.46 &281.5& 80.2 & 0.49 & 0.07* & 0.27 & 0.43 \\
\gemmas &80.0& 113.2 & 0.42 & -0.06* & 0.45 & 0.49 &269.5& 75.5 & 0.43 & -0.03* & 0.44 & 0.49 \\
\midrule
\textbf{Token Instruction Tune}&&&&&&&&&&&&\\
\mbs &88.0& 186.1 & 0.39 & -0.27* & 0.52 & 0.49 &277.8& 667.6 & 0.39 & -0.08* & 0.52 & 0.49 \\
\lms &78.1& 82.8 & 0.40 & -0.06* & 0.50 & 0.51 &276.6& 62.8 & 0.31 & \textbf{0.08}* & 0.66 & 0.64 \\
\gemmas &76.3& 89.8 & 0.34 & -0.08* & 0.59 & 0.60 &229.3& 64.9 & \underline{0.23} & 0.01 & 0.73 & \underline{0.73} \\
\midrule
\textbf{Steering}&&&&&&&&&&&&\\
\mbs &83.4& 133.3 & 0.56 & -0.30* & -0.00 & 0.30 &237.4& 127.7 & 0.55 & -0.15 & 0.01 & 0.28 \\
\lms &84.2& 153.1 & 0.53 & -0.09* & 0.12 & 0.34 &235.1& 183.6 & 0.53 & -0.00 & 0.125 & 0.36 \\
\gemmas &82.1& 138.9 & 0.55 & -0.01 & 0.17 & 0.39 &261.9& 125.0 & 0.56 & 0.03* & 0.12 & 0.35 \\
\midrule
\textbf{\ours}&&&&&&&&&&&&\\
\mbm &87.1& \underline{65.6} & 0.38 & -0.00 & 0.58 & 0.57 &293.6& 63.1 & 0.28 & 0.07* & 0.73 & 0.69 \\
\mbl &77.2& 83.0 & 0.34 & -0.00 & 0.62 & 0.61 &260.3& \textbf{31.4} & \textbf{0.15} & \underline{0.07}* & \textbf{0.85} & \textbf{0.85} \\
\bottomrule
\end{tabular}
}
\end{table*}

\section{Results and Discussion}
\label{sec:results}

\begin{figure*}[t]
    \centering
    \includegraphics[width=1.0\textwidth]{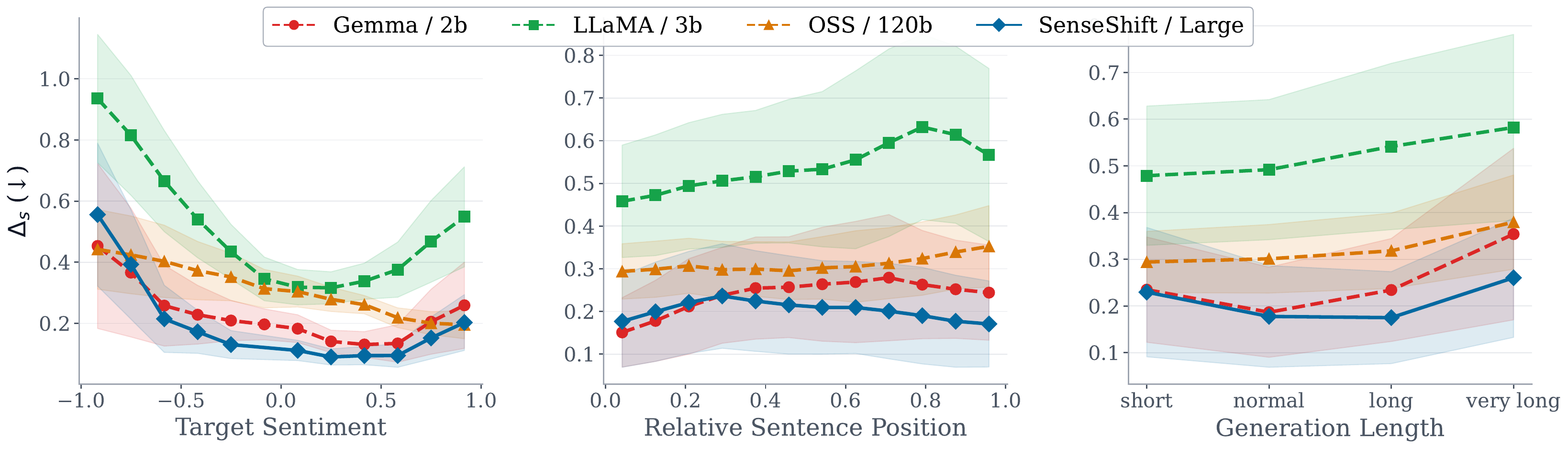}
    \caption{$\Delta_s$ across three evaluation dimensions. Shaded regions indicate variance. \textit{Left:} $\Delta_s$ vs.\ target sentiment.
    \textit{Center:} $\Delta_s$ vs.\ relative sentence position.
    \textit{Right:} $\Delta_s$ vs.\ number of masked tokens.}
    \label{fig:delta_s_analysis}
\end{figure*}

Table~\ref{tab:control_results_id_ood} reports results on the story and review datasets. The top block include prompting baselines that require no training and are domain agnostic. In the table, columns $PPL_{old}$ and $PPL_{new}$ demonstrate the perplexity of the sentences before and after the generation respectively. Note that sentences are selected randomly from the same pool of 2000 evaluation samples hence the selected sentencce perplexity value has variation. More over in general the original $PPL_{old}$ of the review dataset is high due to unconventional human review writing style (e.g., \textit{best best best food that i had}) which shows reveals itself in old sentence perplexity value. The old perplexity of sentences is used as a reference point to give some idea about the distribution of sentence perplexity randomly selected for evaluation and how models new generation perplexity compares with the old sentences. The \textbf{In-Domain Evaluation} block reports models trained and evaluated on the same dataset, while the \textbf{Out-of-Domain Evaluation} block reports models trained on one dataset and evaluated on the other. Note that we use green and red color code to distinguish models trained on story from review. For instance models trained on story are evaluated on story as in-domain while same models evaluated on review as OOD evaluation. The OOD block is provided to allow a fairer comparison with domain agnostic prompting and should be read as robustness check not generalization claim. 

Among prompting baselines, sentiment following improves with scale: \gemmam already has decent sentiment accuracy of 0.55 and \ossl achieves the highest correlation (0.793 on story, 0.74 on review) and accuracy (0.66, 0.64), though at a very high computational cost. Its elevated perplexity relative to \ossm is likely attributable to more complex vocabulary penalized by the GPT-2 scorer. Across prompting baselines, $\Delta_f$ on story are often significantly negative, indicating that prompted sentence generations are typically less contextually well-fitted than the original sentence even-though models have access to the original sentence during prompting unlike other baselines.

Standard instruction tuning improves $\Delta_s$, Corr.\ and Acc.\ over the corresponding prompting baselines, but gains are uneven. \mbs is the most unstable, with perplexity spiking above 400 and better but significant $\Delta_f$ drops, likely reflecting the small size of the model. Token instruction tuning improves even more: \gemmas during In-domain evaluation improves accuracy from 0.49 to 0.75 on stories and 0.49 to 0.67 for reviews as with $\Delta_s$ of 0.21 by injecting \ours sentence-level controllable sentiment tokens. However, most decoder models under this paradigm when evaluated as out-of-domain show significant $\Delta_f$ drops (except for \lms trained on story evaluated on review), suggesting that their generation, while grammatically acceptable, often break contextual fitness. Steering underperforms across all model sizes, with $\Delta_s \approx 0.53$--$0.56$, near-zero or negative correlations, accuracy in the 0.28--0.39 range, and perplexity up to 221, frequently worse than the corresponding prompting baseline even when steering vectors are extracted from the same dataset. With steering only \gemmas shows a marginal improvement, hinting that activation steering might be poorly suited for fine-grained sentiment control at sentence level.

Overall \ours achieves the strongest overall performance. \mbl achieves the lowest perplexity (53.6 on Story, with \mbm at 55.3 on Review) during in-domain evaluation, the smallest sentiment deviation ($\Delta_s$ of 0.20 on Story), and the highest in-domain accuracy on Story (0.78). Crucially, \mbl maintains $\Delta_f$ at or near zero across both domains advocating contextual fitness of the generated sentences. \mbm already delivers strong results relative to all baselines, indicating that \ours scales favorably with model capacity. In the out-of-domain setting, \ours remains competitive with, and on most metrics outperforms, the baselines trained on the matching domain.
% \mbl trained on stories and evaluated on reviews achieves $\Delta_s$ of 0.15, accuracy of 0.85, and perplexity of 31.4.
More importantly, \ours is the only method to maintain near-zero $\Delta_f$ in both the in-domain and out-of-domain blocks, suggesting that encoder conditioning preserves contextual fit while controlling sentiment intensity robust to change of training source.

\paragraph{Sentiment Following Analysis.}
Next we analyze sentiment following capabilities of competing baselines by examining how $\Delta_s$ is influenced by: (1) target sentiment specified by the user, (2) relative sentence position $\left(\frac{\text{sentence index}}{\text{total number of sentences}}\right)$, and (3) generation length as words: short ($<5$), normal ($5 \leq l < 10$), long ($10 \leq l < 20$), and very-long ($\geq 20$). We compare \mbl with competitive baselines namely prompting \ossl and token instruction tuning of \lms, and \gemmas. 

Figure~\ref{fig:delta_s_analysis} visualizes these analysis. On left, all models demonstrate higher $\Delta_s$ when targeting strongly negative sentiment values, whereas performance improves then slightly deteriorates toward highly positive targets. Exceptionally, \ossl demonstrates a monotonic improvement as the target sentiment becomes more positive while performing slightly better than \mbl at extreme sentiments. \lms, \gemmas and \mbl all exhibit their lowest $\Delta_s$ around neutral zone, showing difficulty in producing highly polarized outputs despite strong overall controllability. We attribute this behavior to sentiment distribution of the training corpus for trained models. The story dataset sentence sentiments are heavily skewed toward neutral and positive sentiment, which likely biases models toward conservative sentiment shifts (see Table~\ref{tab:dataset_stats}). 

With respect to sentence position, all decoder models exhibit a mild increase in $\Delta_s$ when the target sentence occurs later in the story. This suggests that models benefit less from richer preceding context when performing sentiment-controlled generation. This behavior is particularly different for \mbl, which maintain the least $\Delta_s$, however exhibits higher $\Delta_s$ on earlier sentences but improves steadily toward later positions. We hypothesize that encoder-based architectures rely more heavily on contextual buildup, making early-sentence editing inherently more challenging due to limited narrative grounding.

Finally, we analyze the effect of generation length. \lms and \ossl show increased $\Delta_s$ as generation length grows, while \mbl overall has the least $\Delta_s$ and again demonstrates a different trend: $\Delta_s$ decreases for normal and long generations before slightly increasing again. \gemmas also shows improved control for normal length generations before degrading on longer sentences.\footnote{Joint analysis of \ours is provided in Section~\ref{app:sec:extra_results}.} In summary, our analysis suggests that \ours maintains a robust $\Delta_s$ advantage over baselines independent of target sentiment, sentence position, and generation length, indicating that the gap reflects a genuine difficulty in maintaining sentiment coherence rather than an artifact of any single evaluation condition.

% \begin{table}[h]
% \centering
% \small
% \caption{Users preference ($\%$) of sentences generated by \mbl-story vs \ossl. Rows correspond to evaluation criteria: contextual fitness, grammatical correctness, and sentiment following. Similarity to original sentence is for diagnostics and is excluded from overall score.}
% \resizebox{\columnwidth}{!}{%
% \label{tab:human_eval}
% \begin{tabular}{lcccc}
% \toprule
% \textbf{Criterion} & \textbf{\ossl} & \textbf{\ours} & \textbf{Both} &\textbf{Neither} \\
% \midrule
% Fitness       & 46.2 & 17.2 & 31.2 &5.4\\
% Fluency & 23.7 & 10.8 & 63.4&2.2 \\
% Sentiment & 55.9 & 14.0 & 24.7& 5.4\\
% \textbf{Overall}      & 41.9 & 14.0 & 39.8& 4.3\\
% \midrule
% Similarity & 78.9 & 9.2 & 6.1& 5.7\\
% \bottomrule
% \end{tabular}
% }
% \end{table}

\begin{table}[hbt]
\centering
\small
\caption{Users preference ($\%$) of sentences generated by \mbl-story vs \gemmas trained with Token-Instruction-Tuning. Rows correspond to evaluation criteria: contextual fitness, grammatical correctness, and sentiment following. }
% Similarity to original sentence is for diagnostics and is excluded from overall score.}
\resizebox{\columnwidth}{!}{%
\label{tab:human_eval}
\begin{tabular}{lcccc}
\toprule
\textbf{Criterion} & \textbf{\gemmas} & \textbf{\ours} & \textbf{Both} &\textbf{Neither} \\
\midrule
Fitness       &22.1& 23.0 & 46.9 &8.0\\
Fluency & 6.2 &  6.2& 85.0&2.7 \\
Sentiment & 28.3 & 31.0 & 28.3& 12.4\\
\midrule
\textbf{Overall}      & 18.9 & 20.1& 53.4& 7.7\\
% \midrule
% Similarity & 6.2 & 9.2 & 18.8& 5.7\\
\bottomrule
\end{tabular}
}
\end{table} 

\paragraph{Human Evaluation.}
Table~\ref{tab:human_eval} summarizes human preference results across three criteria: contextual fit, grammatical fluency, and sentiment following. Notably, \mbl achieves these results with only 395M parameters, compared to \gemmas (2B parameters) a 5$\times$ reduction in parameters (described in Section\ref{subsec:evalation_metric}).

On fluency and contextual fit, both models perform comparably. Annotators rated the outputs as equally fluent in 85.0\% of samples and equally well-fitting to the surrounding context in 46.9\% of cases, indicating that sentiment control does not substantially degrade linguistic quality. Among the remaining contextual-fit judgments, \mbl is preferred at a slightly higher rate to \gemmas (23.0\% vs.\ 22.1\%), showing that the smaller encoder-based model matches a substantially larger generative model on contextual integration.

Differences are more pronounced for sentiment following. Annotators judged \mbl sentiment satisfactory in 31.0\% of examples versus 28.3\% for \gemmas, selected both models in 28.3\%, and preferred neither model in 12.4\% of cases. This indicates that \mbl achieves stronger sentiment alignment while preserving contextual coherence.

Overall, annotators rated \mbl as equivalent to or better than \gemmas in 73.5\% of evaluated samples. Achieving this with a fraction of the parameters suggests that encoder-centric approaches are a strong and competitive alternative for fine-grained sentiment-aware controllable text generation, particularly in settings requiring simultaneous preservation of context and precise sentiment manipulation.

\section{Conclusion}
\label{sec:conclusion}

This paper introduced \ours, a lightweight generative framework base on encoder language models for fine-grained sentiment-aware in-text generation. \ours utilizes quantized sentence level sentiment token, to condition mask prediction of encoder models. Using our modfied sequential mask prediction strategy \ours is able to produce coherent text following the fine-grained sentiment signal. Through extensive in-domain and out-of-domain evaluation on story and review generation of two encoder models in two sizes we demonstrate that \ours can generate sentences at various positions using fine-grained sentiment control signals, while preserving overall fluency, contextual fitness, and sentiment adherency compare to fine-tuned and prompting autoregressive models. Furthermore, our results indicate that \mbl performs robust in out-of-domain scenarios, allowing stable performance for in-text generation task.

\section{Limitations and Ethical Consideration}
\label{sec:limitations}

\paragraph{Sentiment Analysis.}
 Our framework relies on VADER~\cite{vader2016} for all sentiment scoring, both during training signal construction and at inference time, where sentiment must be evaluated repeatedly over the full text to guide and assess each rewrite. This design choice priorities speed: More accurate neural sentiment models (e.g., RoBERTa~\cite{barbieri-etal-2020-tweeteval}) would substantially increase latency per generation, damaging a main contribution of the work, i.e., providing \textit{lightweight} framework. VADER's deterministic nature, however, limits sensitivity to implicit or contextually nuanced sentiment, and reported $\Delta_s$ values should be interpreted with this constraint in mind. Access to human level sentence sentiment or replacing VADER with a lightweight but more accurate sentiment scorer is a promising direction for future work. 
 
 Furthermore, sentiment is an inherently subjective construct: human annotators frequently disagree on the polarity and intensity of the same sentence, which places an upper bound on how precisely any model including \ours can be evaluated against a "true" sentiment label.

\paragraph{Training Corpora and Sentiment Distribution.} Our models are trained on  TinyStories and a review corpus, both of which are narrow in domain. The choice of dataset stems from the task which supposedly requires changing dynamics of the sentiment (partly negative, partly positive) which can be found in long text generation. These constraints prevent us from making strong claims about true generalization to unseen domains such as news, social media, or formal writing. A further challenge is the sentiment distribution of both corpora: Extreme sentiment values appear very infrequently, meaning the model sees comparatively few training examples of strongly positive or strongly negative targets, with negative ones being significantly less observed. This imbalance makes fine-grained control at the tails of the sentiment range harder, and likely contributes to the elevated $\Delta_s$ we observe at $s = -1.0$. We expect that curating higher-quality, sentiment-balanced training data, particularly with more uniform coverage of extreme values would meaningfully improve performance, and we leave this exploration to future work.

During editing we observed that the errors of the model with respect to target sentiment also is effected by the sentiment of the sentences surrounding it where the more positive the context surrounding , the more edits models required to change the sentiment to opposite polarity. Due to the constrained scope we did not delve into this effect and leave this analysis to future works. 

\paragraph{Baseline Selection and Compute Constraints.} Baselines were selected to enable quantifiable, reproducible comparison across prompting, instruction tuning, and activation steering paradigms with a special version of instruction tuning with our sentence level quantized tokenization of sentiment. The choice of baselines was based on compatibility with our method, and comparison between decoder and encoder model architectures. The specific models chosen reflect resource constraints: All experiments were conducted under a fixed compute budget, which precluded evaluation of the largest available open-weight models. Results for prompting-based baselines should therefore be read as representative rather than exhaustive, and stronger frontier models may narrow the observed performance gap.

Our model is designed to edit text toward an arbitrary target sentiment intensity, including strongly negative values. This capability carries clear misuse potential. Because edits are designed to be fluent and contextually coherent, modified outputs may be difficult to distinguish from originals without provenance tracking. We strongly recommend that any deployment of sentiment controllable systems be accompanied by appropriate safeguards, including use-case restrictions, output disclosure to end users, and, where feasible, watermarking or audit logging. Responsible use of this technology is the obligation of both developers and practitioners who build upon it.

\bibstyle{acl_natbib}
\nocite{Gusfield:97}
\bibliography{custom}

\appendix

\section{Appendix}
\label{sec:appendix}

\subsection{Problem Statement and Motivation}
\label{sec:app:gpt_story_generation}

\begin{figure}[t]
    \centering
    \includegraphics[width=0.5\textwidth]{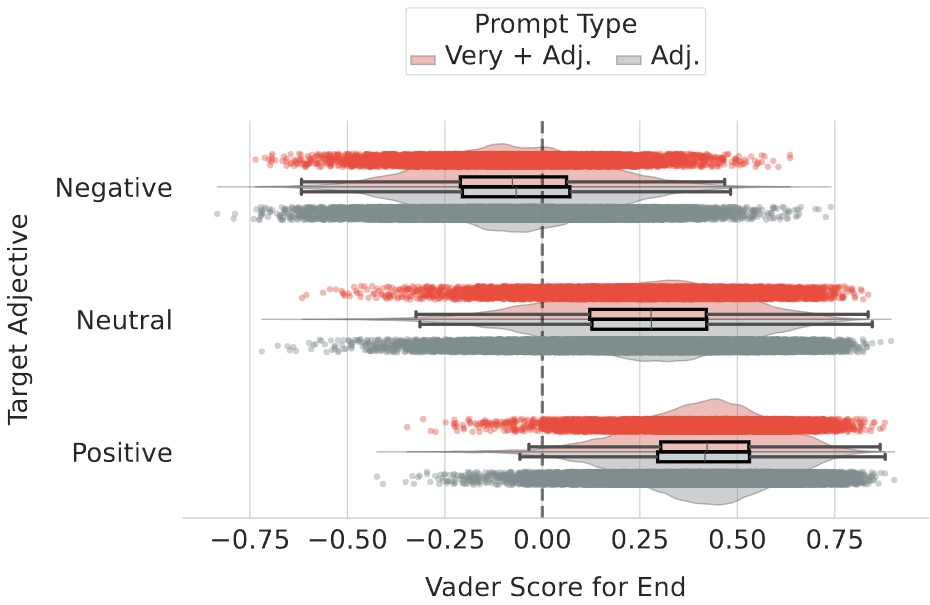}
    \caption{Sentiment of story ending written by \gptmini for Negative, Positive, and Neutral ending with and without intensifying adverb "Very" in the prompt.}
    \label{fig:sentiment_analysis}
\end{figure}

Further to related research supporting LLMs challenges of controllability through prompting we also did a small experiment to further show that models are incapable of following prompt-based controlling strategies. We directed \gpts model to provide stories with three distinct endings namely Positive, Negative, and Neutral with/without intensifying adverb "Very". We passed the endings provided by \gpts to VADER sentiment analyzer to score the sentiment of the stories. Figure~\ref{fig:sentiment_analysis} shows the result of this experiment. As it can be seen from the figure there are three major problems with the provided endings: (1) weak separation between adjective and intensified-adjective conditions, indicating limited intensity control~\cite{samuel_cie_2025}; (2) a distributional shift toward more positive sentiment, even under negative targets~\cite{chen-etal-2025-labor}; and (3) compressed score ranges that rarely reach sentiment extremes. These findings are consistent with earlier reports that LLMs struggle with fine-grained controllability and strict prompt constraints \citep{sun_evaluating_2023,lu_bounding_2023}. Together, this motivates moving beyond prompt-only control and focus on architectural changes and controllable frameworks.

\subsection{Baselines}
\label{sec:app:baselines}

In This section we provide more technical detail on our baselines namely, Prompting, Instruction Tuning, Token Instruction Tuning and Steering Activation.

\subsubsection{Prompts}
\label{subsec:app:prompts}
When evaluating decoder models the choice of prompt is of great importance, slight changes in the prompt might lead to variations in the outcome of the model which further motivates our work. However, In order to have a robust evaluation we modified and re-evaluated the decoder models iteratively to come-up with the best instructions that led to the most accurate outcome in both response of the model and accuracy in following the sentiment. In this section we provide specific prompts used for each of the evaluation. Note that prompting variation is only important when evaluating off-the-shelf models and for instruction tuning, token instruction tuning the choice of prompt is not strongly related as the model will be evaluated with the same prompt to ensure a fair comparison with \ours. 

Prompting: For prompting we observed that masking the target sentence and asking to generate a sentence that fits best to that location is not performing accurately hence the prompt provides the full text while asking model to replace the existing sentence in text with a new sentence of another sentiment. Below we can find the specific prompt for every setup.

\tcbset{
    orangebox/.style={
        colback=orange!15,   % background color (light orange)
        colframe=orange!75,  % border color
        boxrule=0.8pt,       % border thickness
        arc=4pt,             % rounded corners
        left=4pt, right=4pt, top=4pt, bottom=4pt,
        fonttitle=\bfseries,
    }
}

% Then in your appendix:
\begin{tcolorbox}[orangebox, title=Prompting]
Consider that every sentence has a sentiment where -1 is the most negative and 1 is the most positive sentiment. Given this text:

\texttt{\{FULL\_TEXT\}}

Rewrite {TARGET\_SENTENCE} to have a sentiment value of {TARGET\_SENTIMENT} but still make sense to overall text. Write only the sentence and nothing else.

\end{tcolorbox}

Instruction Tuning: For instruction tuning to work the model requires the sentence to be predicted hence having the sentence in the text was not desirable and considered as leakage of information, Hence we modified the instruction and included {[MISSING\_SENTENCE]} as masked version of the sentence and asked the models to write the missing sentence to have the specific value.

% Consider that every sentence has a sentiment where -1 is the most negative and 1 is the most positive sentiment. Given this text: \n{MASKED\_TEXT}\n" Rewrite {[MISSING\_SENTENCE]} to have a sentiment value of {TARGET\_SENTIMENT} but still make sense to overall text. Write only the sentence and nothing else.

\tcbset{
    orangebox/.style={
        colback=orange!15,   % background color (light orange)
        colframe=orange!75,  % border color
        boxrule=0.8pt,       % border thickness
        arc=4pt,             % rounded corners
        left=4pt, right=4pt, top=4pt, bottom=4pt,
        fonttitle=\bfseries,
    }
}

\begin{tcolorbox}[orangebox, title=Instruction Tuning]
Consider that every sentence has a sentiment where -1 is the most negative and 1 is the most positive sentiment. Given this text:

\texttt{\{MASKED\_TEXT\}}

Rewrite {[MISSING\_SENTENCE]} to have a sentiment value of {TARGET\_SENTIMENT} but still make sense to overall text. Write only the sentence and nothing else. 
\end{tcolorbox}

Token Instruction Tuning: The difference between this variation and the normal instruction tuning is the introduction of new controlling tokens to the model just as \ours is trained on. We equiped decoder models with \ours sentiment tokens $[\sigma_i]$ as special untrainable tokens as well as added {[MISSING\_SENTENCE]} as a new special token totalling 22 new tokens. in this version of the prompt each sentence has the control signals attached at the beginning of every sentence to give them context just like \ours:

% Consider that every sentence has a sentiment where -1 is the most negative and 1 is the most positive sentiment. Given this text: \n{MASKED_TEXT_WITH_TOKENS}\n" Rewrite {[MISSING_SENTENCE]} to have a sentiment value of {$[\sigma_i]$} but still make sense to overall text. Write only the sentence and nothing else. {$[\sigma_i]$}

 % put this in your preamble

% Define a custom orange box style
\tcbset{
    orangebox/.style={
        colback=orange!15,   % background color (light orange)
        colframe=orange!75,  % border color
        boxrule=0.8pt,       % border thickness
        arc=4pt,             % rounded corners
        left=4pt, right=4pt, top=4pt, bottom=4pt,
        fonttitle=\bfseries,
    }
}

\begin{tcolorbox}[orangebox, title=Token Instruction Tuning]
Consider that every sentence has a sentiment where -1 is the most negative and 1 is the most positive sentiment. Given this text: 

\texttt{\{MASKED\_TEXT\_WITH\_SENTIMENT\_TOKENS\}}  

Rewrite \texttt{\{[MISSING\_SENTENCE]\}} to have a sentiment value of $\{[\sigma_i]\}$ but still make sense to overall text. Write only the sentence and nothing else. $\{[\sigma_i]\}$
\end{tcolorbox}

\subsubsection{Activation Steering}
\label{subsec:app:steering}

Over the past few years, activation steering has emerged as an effective technique for controlling generated text using contrastive samples. These samples are typically constructed from the original dataset as minimally differing pairs, where small, targeted changes (e.g., sentiment polarity) induce measurable shifts in model activations. Although this approach does not require additional model training, its effectiveness depends critically on the quality and construction of the contrastive pairs.

In our setup, we construct contrastive pairs by selecting sentences with extreme sentiment values (i.e., $|\sigma_{s_i}|>0.9$). We then use the steering vector library~\footnote{https://github.com/steering-vectors/steering-vectors} to extract activation directions from these pairs. Here, “training” refers not to updating model parameters, but to computing steering vectors that capture the latent direction corresponding to the desired attribute shift. For each dataset, we sample $10k$ contrastive pairs from the pool of possible combinations to estimate these directions.

At inference time, the steering strength is controlled via a scaling factor (the multiplier), which determines the extent to which the model output is shifted toward the target attribute. While the original method is primarily designed for categorical control, it also suggests the potential for continuous modulation. Following this insight, we adapt the multiplier dynamically based on the target sentiment value, enabling more fine-grained, continuous control over the generated text.

\subsubsection{Dataset Statistics and Training Hyper-Parameters}
\label{sec:app:training}

During pre-processing we removed major samples in the dataset which were not suitable for our task (e.g., reviews with one sentence). In Table~\ref{app:tab:hyperparameters} we report the statistics of the training data and hyperparameters used for model training.
\begin{table}
\centering
\caption{Dataset statistics for the train splits. Sentiment scores are in $[-1, 1]$.}
\label{tab:dataset_stats}
\resizebox{\columnwidth}{!}{%
\begin{tabular}{l|c|c}
\toprule
\textbf{Statistics} & \textbf{TinyStories} & \textbf{Yelp} \\
\midrule
Num Samples & 4439352 & 252085 \\
Positive (>0.2) & 53.7\% & 41.0\% \\
Negative (<-0.2) & 0.3\% & 4.3\% \\
Neutral & 46.0\% & 54.8\% \\
\midrule
\multicolumn{3}{c}{\textbf{Text}}\\ 
\midrule
Avg Sentences & 14.2 ± 7.77 & 6.4 ± 4.86 \\
Median Sentences & 13. & 5 \\
Min/Max Sentences & 2 / 195 & 2 / 86 \\
Avg Words & 152.90 ± 53.52 & 90.28 ± 80.34 \\
Median Words & 143.00 & 68.00 \\
Min/Max Words & 9 / 902 & 1 / 1004 \\
Avg VADER& 0.21 ± 0.14 & 0.15 ± 0.20 \\
Median VADER& 0.21 & 0.14 \\
Min/Max VADER& -0.68 / 0.92 & -0.91 / 0.96 \\
Avg RoBERTa& 0.27 ± 0.22 & -0.0206 ± 0.64 \\
\midrule
\multicolumn{3}{c}{\textbf{Sentences}} \\ 
\midrule
Avg Words & 10.7 ± 5.36 & 13.9 ± 9.61 \\
Median Words & 10 & 12 \\
Min/Max Words & 1 / 702 & 1 / 550 \\
Avg VADER sent. & 0.1978 ± 0.38 & 0.13 ± 0.37 \\
Min/Max VADER & -1.0 / 1.0 & -0.99 / 0.99 \\
Avg RoBERTa sent. & 0.25 / 0.52 & -0.078 / 0.94 \\
Min/Max RoBERTa& -1.0 / 1.0 & -0.99 / 0.99 \\
\bottomrule
\end{tabular}
}
\end{table}
For our models we have not used any PEFT method and all models' full number of parameters are tuned for the task to ensure that none of the results are effected by the number of parameters that are trained. %Table~\ref{app:tab:hyperparameters} summarizes the training hyperparameters used for the training of the models. 

\begin{table*}[!ht]
\centering
\small
\setlength{\tabcolsep}{5pt}
\caption{Default hyperparameters used across different training paradigms for controllability. * value for Qwen is 1024 to ensure finalizing thinking process.}
\begin{tabular}{ll|cccc}

\toprule
\textbf{State} & 
\textbf{Hyperparameter} & \textbf{Prompt/Steer} &
\textbf{Inst-Tune} & \textbf{Token-Inst-Tune} & \textbf{\ours} \\
\midrule
\multirow{7}{*}{Training}  
& Learning rate (lr)     & --     & 1e-5  & 1e-5  & 1e-5 \\ 
& Batch size (bs)        & --     & 32    & 32    & 50   \\ 
& Dropout (dp)           & --     & 0.0   & 0.0   & 0.0  \\ 
& Warmup steps           & --     & 500   & 500   & 500 \\ 
& Epochs                 & --     & 30     & 30     & 50 \\ 
& Weight decay           & --     & 0.01  & 0.01  & 0.01 \\ 
& Scheduler           & --     & Linear  & Linear  & Linear \\ 
& Regime           & --     & Causal  & Causal  & MLM \\ 

% \cline{1-6}
\midrule

\multirow{4}{*}{Generation}  
& Temperature (T)        & 0.7    & 0.7   & 0.7   & 0.9  \\ 
& Top-p                  & 0.9    & 0.9   & 0.9   & 0.9  \\ 
& Repetition penalty (rp)& 1.1    & 1.1   & 1.1   & -  \\ 
& Max tokens             & 128*  & 128   & 128   & 30  \\ 

% \cline{1-6}
\bottomrule
\end{tabular}
\label{app:tab:hyperparameters}
\end{table*}

\subsection{Iterative Infilling Versus One Shot Prediction}
\label{sec:app:iter_v_noniter}

\begin{table*}[!ht]
\caption{Comparison between Iterative Mask Infilling and Whole Mask prediction}
\centering
\small
\setlength{\tabcolsep}{5pt}
\resizebox{1.0\textwidth}{!}{%
\begin{tabular}{lc|ccccc|ccccc}

\toprule

  \multirow{2}{*}{\textbf{Model}} &
  \multirow{2}{*}{\textbf{Train}} &
  \multicolumn{5}{c|}{\textbf{Story Generation}} &
  \multicolumn{5}{c}{\textbf{Review Generation}} \\
  
& & \textbf{Ppl. $\downarrow$} &\textbf{$\Delta_s$ } $\downarrow$ & $\Delta_f$ $\uparrow$ & \textbf{Corr.} $\uparrow$ & \textbf{Acc.} $\uparrow$ &\textbf{Ppl. $\downarrow$} & \textbf{$\Delta_s$} $\downarrow$ & $\Delta_f$ $\uparrow$ & \textbf{Corr.} $\uparrow$ &\textbf {Acc.} $\uparrow$ \\

\midrule

\textbf{Whole Mask Prediction}&&&&&&&&&&\\\
\multirow{2}{*}{\mbm}  & Story & 596.8 & 0.42 & -0.01 & 0.54 & 0.44 & 303.1 & 0.44 & -0.02* & 0.48 & 0.38 \\
                       & Review & 482.9 & 0.50 & -0.04* & 0.27 & 0.31 & 325.2 & 0.43 & -0.01 & 0.55 & 0.44 \\

\multirow{2}{*}{\mbl} & Story & 869.9 & 0.4 & -0.039* & 0.55 & 0.5 & 571.3 & 0.43 & 0.01 & 0.48 & 0.39  \\
                      & Review & 426.3 & 0.50 & -0.041* & 0.29 & 0.32 & 288.8 & 0.44 & -0.013 & 0.51 & 0.41  \\

\midrule 
\textbf{Iterative Mask infilling}&&&&&&&&&&\\\
\multirow{2}{*}{\mbm}  & Story & 91.6 & \underline{0.26} & \underline{-0.00} & \underline{0.77} & \underline{0.69} & 63.1 & 0.28 & \textbf{0.07}* & \underline{0.73} & \underline{0.69} \\
                       & Review & \underline{65.6} & 0.38 & -0.00 & 0.58 & 0.57 & \underline{55.3} & 0.29 & 0.06* & 0.71 & 0.66 \\

\multirow{2}{*}{\mbl} & Story & \textbf{53.6} & \textbf{0.20} & -0.00 & \textbf{0.79} & \textbf{0.78} & \textbf{31.4} & \textbf{0.15} & \underline{0.07}* & \textbf{0.85} & \textbf{0.85} \\
                      & Review & 83.0 & 0.34 & \textbf{-0.00} & 0.62 & 0.61 & 57.5 & \underline{0.26} & 0.04* & 0.71 & 0.69 \\
\bottomrule
\end{tabular}
}
\label{app:tab:it_vs_nonit}
\end{table*}

Mask prediction was not used for long text generation due to their one-shot prediction of all masks which result in several repetitive high probability tokens appearing in different parts of the text without rational connection. Here we present an ablation study of removing infilling mask prediction from the model to show the effectiveness of the technique. 

Table~\ref{app:tab:it_vs_nonit} compares our iterative mask infilling strategy against a whole mask prediction using \ours, in which all masked tokens are predicted in a single forward pass. The results strongly favour the iterative approach across every metric and configuration. Perplexity improvements are the most striking: iterative infilling reduces PPL by a factor of 5--10$\times$ across all model sizes and training domains (e.g., \mbl Story drops from 869.9 to 53.6), indicating that generating masked spans token-by-token produces substantially more fluent text than predicting the entire span at once. Sentiment control follows the same pattern: $\Delta_s$ is consistently lower under iterative infilling, with reductions of 0.15--0.25 across configurations, while $\Delta_f$ values remain near zero compared to the small but consistent negative drift observed under whole mask prediction. These results suggest that whole mask prediction suffers from exposure bias where the model must generate a coherent span without access to its own intermediate outputs whereas iterative infilling conditions each token on the previously generated context, allowing sentiment and fluency constraints to be satisfied progressively rather than in a single unconstrained step giving advantage of decoder models to encoder ones.

\subsection{Human Evaluation Protocol}
\label{bsec:app:human_eval}

To ensure that our results are comparable with human understanding of fitness and sentiment we compared \ours rewrites with parallel rewrites of closest comparable competing decoder-based model namely \gemmas. Both models are trained on story and \gemmas is selected for human evaluation because of its competetive performance in automatic evaluation. Furthermore \gemmas is the closest comparable baseline when it comes to evaluation while it goes through the same sentiment token training as \mbl and in contrast to prompting method which the original sentence is visible to the model \gemmas does not have access to the original sentence. 

Our human evaluation is a user preference study where users are asked to select which model they think generated a higher quality sentence given a document. We developed a simple annotation platform and selected 100 samples from stories (n=55) and reviews (n=45) where models both models rewrite for the exact same sentence and the exact same sentiment. We shared the annotation link with 8 annotators volunteered to give their preference on the rewrites. We did not collect their demographic information nor collected any other sensitive information rather than labels from the annotators. We provided them full document context (review or story), the original sentence which was targeted for replacement, and two anonymized candidate generated sentences which one produced by \gemmam trained with token instruction tuning and one by \mbl. We display the sentences in randomized order to control for position bias of annotators and also to protect model identities. For each pair, annotators respond to three questions:  
\begin{itemize}
    \item \textbf{Contextual Fit.} Which rewrite fits better into the surrounding text? Judge whether it matches the tone, vocabulary, register, and stylistic quirks of the full passage. If the original writing is formal, poetic, blunt, or otherwise distinctive, a good rewrite preserves those features and sits well inside the position to produce a coherent text.

    \item \textbf{Grammar and Fluency.} Ignoring the surrounding text, which sentence is more grammatically correct and naturally written as a standalone sentence? Annotators focus purely on syntax, word choice, and fluency.

    \item \textbf{Sentiment Match.} Which rewrite better achieves the target sentiment, specified on a scale from $-1$ (very negative) to $+1$ (very positive)? Annotators judge how well each sentence's emotional tone matches the indicated target value.

    % \item \textbf{Similarity to Original.} Which rewrite is closer to the original sentence in meaning, structure, and phrasing? This question is included as a \textit{diagnostic metric} rather than a quality measure: since \ossl receives the original sentence as part of its prompt and is asked to rewrite it, a high similarity score reflects lexical carryover from the source rather than generative quality. This metric is therefore reported separately and excluded from the quality-based overall score.
\end{itemize}

For each criterion, annotators select one of four options: \textbf{A}, \textbf{B}, \textbf{Both} (both outputs are equally good), or \textbf{Neither} (neither output is satisfactory) where A and B represent \gemmas and \mbl-story in random order. We report average preference of annotators in Table~\ref{tab:human_eval} across the three quality criteria. Tables~\ref{tab:human_eval_story} and~\ref{tab:human_eval_review} report the results in detail for each domain.

\begin{table}[h]
\centering
\small
\caption{Human preference evaluation on the \textbf{Story} domain 
(n=55 sentence pairs). The specific model used for evaluation is \mbl trained on stories. Similarity to Original is reported as a diagnostic metric and excluded from the Overall score.}
\label{tab:human_eval_story}
\resizebox{\columnwidth}{!}{%
\begin{tabular}{lcccc}
\toprule
\textbf{Criterion} & \textbf{\gemmas} & \textbf{\ours} & 
\textbf{Both} & \textbf{Neither} \\
\midrule
Fitness         & 17.2 & 23.4 & 54.7 & 4.7 \\
Grammar & 4.7 & 4.7 & 89.1 & 1.6 \\
Sentiment       & 29.7 & 28.1 & 31.2 & 10.9 \\
\midrule
\textbf{Overall}        & 17.2 & 18.8 & 58.3 & 5.7 \\
\bottomrule
\end{tabular}
}
\end{table}

\begin{table}[h]
\centering
\small
\caption{Human preference evaluation on the \textbf{Review} domain (OOD n=45 sentence pairs).  The specific model used for evaluation is \mbl trained on stories. Similarity to Original is reported as a diagnostic metric and excluded from the Overall score.}
\label{tab:human_eval_review}
\resizebox{\columnwidth}{!}{%
\begin{tabular}{lcccc}
\toprule
\textbf{Criterion} & \textbf{\gemmas} & \textbf{\ours} & 
\textbf{Both} & \textbf{Neither} \\
\midrule
Fitness         & 27.7 &  23.4 & 36.2 & 12.8 \\
Grammar & 8.5 &  8.5 & 78.7 & 4.3 \\
Sentiment  & 25.5 &  36.2 & 23.4 & 14.9 \\
\midrule
\textbf{Overall}        & 20.6 &  22.7 & 46.1 & 10.6 \\
\bottomrule
\end{tabular}
}
\end{table}

\subsection{Joint Analysis of $\Delta_s$}
\label{app:sec:extra_results}

\begin{figure*}[!htbp]
    \centering
    \includegraphics[width=\textwidth]{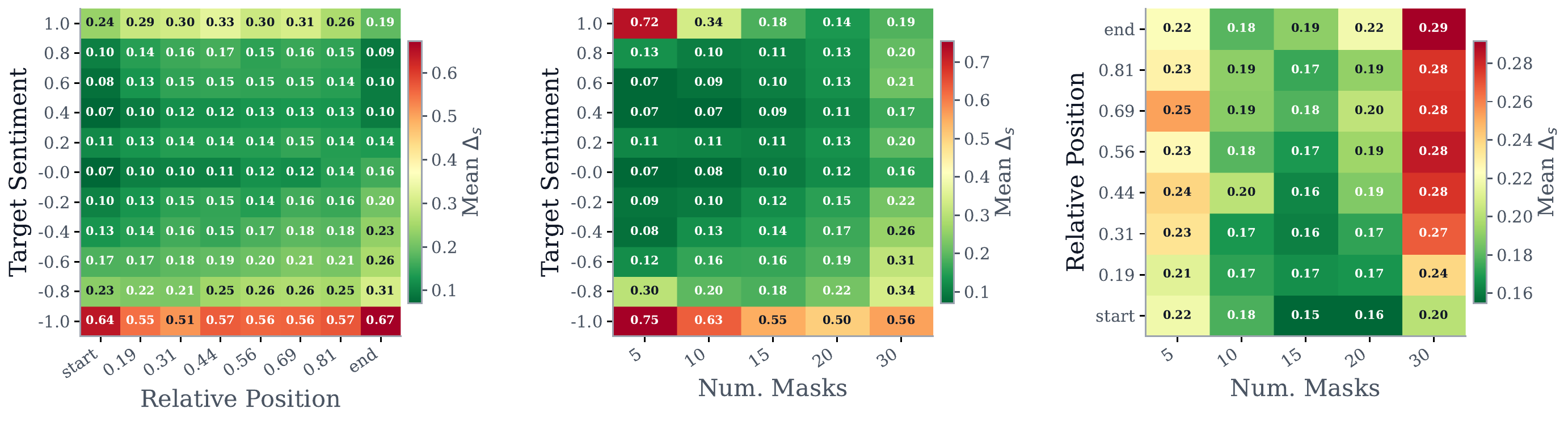}
    \caption{Mean $\Delta_s$ as a joint function for the \mbl-story.
    \textit{Left:} Target sentiment vs. relative sentence position.
    \textit{Center:} Target sentiment vs. number of masked tokens.
    \textit{Right:} Relative sentence position vs. number of masked tokens.
    Lower values (green) indicate better sentiment control.}
    \label{fig:delta_s_heatmaps}
\end{figure*}

Figure~\ref{fig:delta_s_heatmaps} extends the univariate analysis of Section~\ref{sec:results} by examining $\Delta_s$ across pairs of evaluation dimensions for the \textit{large-story} model.

Strongly negative target sentiment ($s{=}{-1.0}$) dominates error across all positions, peaking at $\Delta_s{=}0.66$ at late document positions where accumulated context compounds the difficulty. Across non-extreme sentiment values, sentence position has comparatively little effect, suggesting position-dependent degradation is largely confined to the sentiment extremes.

The interaction between large mask counts and strongly negative targets is the most pronounced pattern across all three heatmaps, reaching a global maximum of $\Delta_s{=}0.72$. At neutral-to-positive targets, increasing mask size has only a modest effect, indicating that additional generation freedom is primarily problematic when the target register is already difficult to maintain.

Error increases moderately toward late positions and large mask counts, though the interaction between the two is approximately additive. Notably, mid-document positions (0.44--0.69) yield the lowest $\Delta_s$, suggesting that a moderate amount of preceding context aids sentiment following more than editing at the document start.

\subsection{Qualitative Examples}
\label{sec:app:example}

\begin{figure*}[!htbp]
    \centering
    \includegraphics[width=\textwidth]{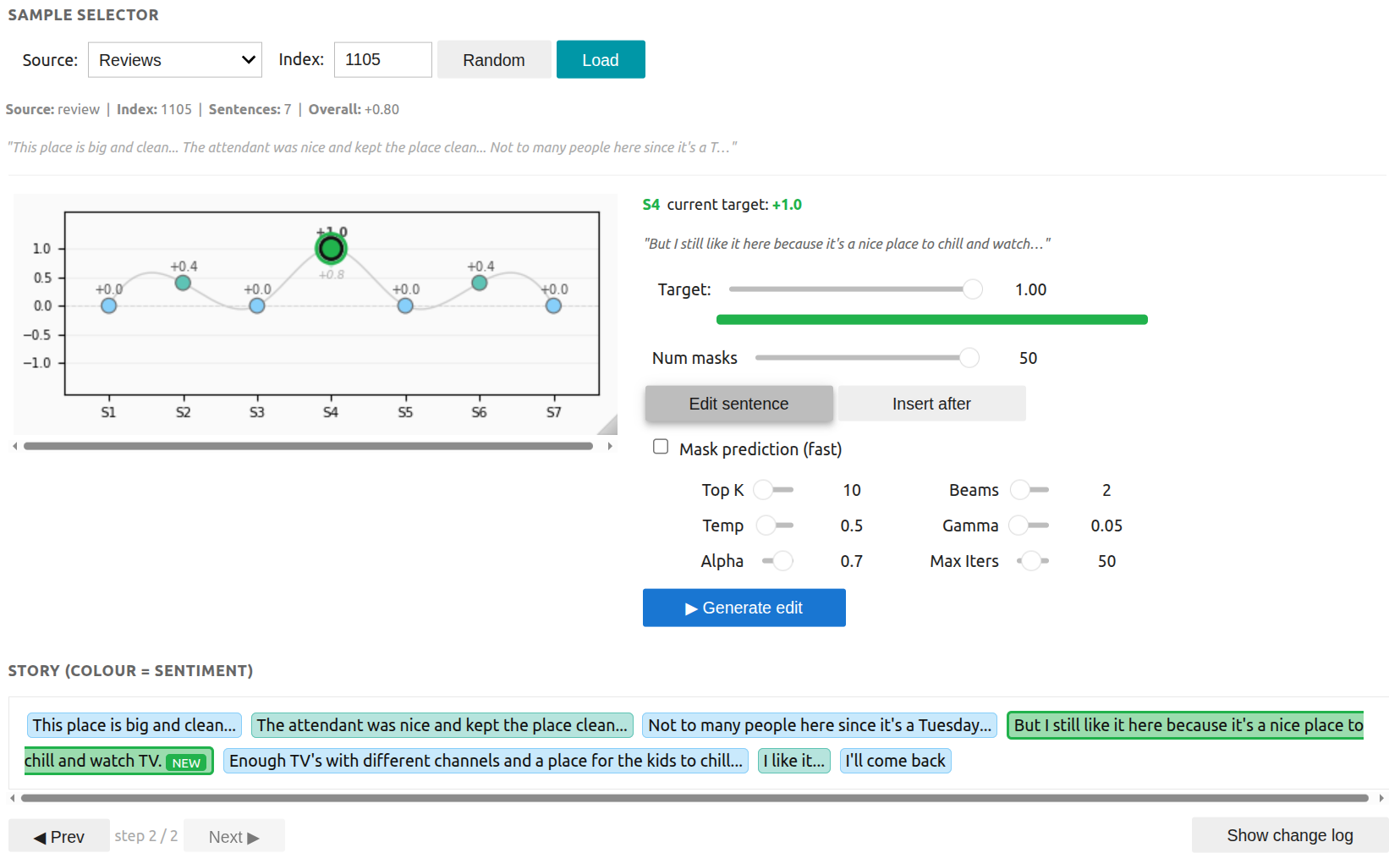}
    \caption{Interactive demo of \ours. Users can choose or randomly select data from both dataset and edit different sentences with their desired sentiment intensity}
    \label{app:fig:demo}
\end{figure*}

To demonstrate the model’s capabilities, we provide a demo program that allows users to inspect model behavior on arbitrary input text. Figure~\ref{app:fig:demo} visualizes the interactive demo. We present qualitative examples of iterative sentiment editing applied to a story (in-distribution) and a hotel review (out-of-distribution), both edited using \mbl trained on the story dataset.

\begin{figure*}[!htbp]
    \centering
    \begin{subfigure}{\textwidth}
        \centering
        \includegraphics[width=0.5\columnwidth]{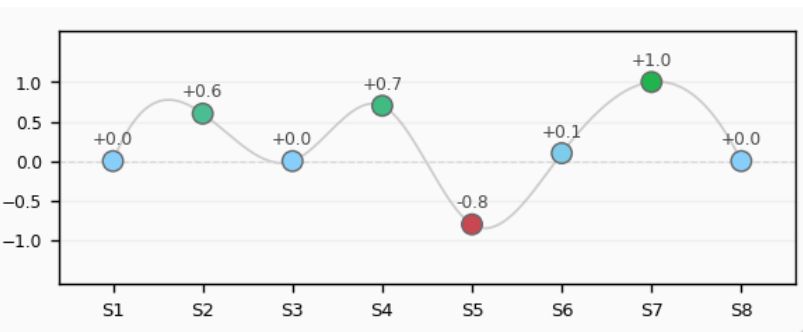}\\[2pt]
        \includegraphics[width=\textwidth]{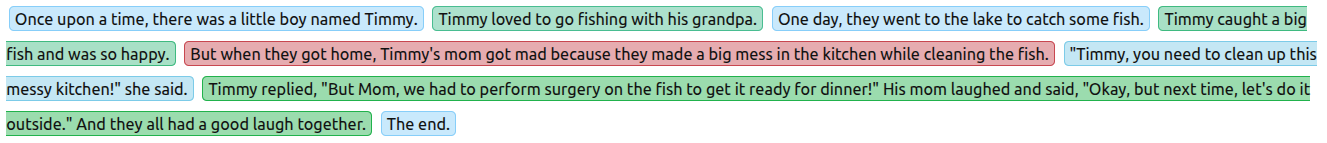}
        \caption{Original story and its sentiment arc.}
        \label{app:fig:story_original}
    \end{subfigure}\\[4pt]
    \begin{subfigure}{\textwidth}
        \centering
        \includegraphics[width=0.5\columnwidth]{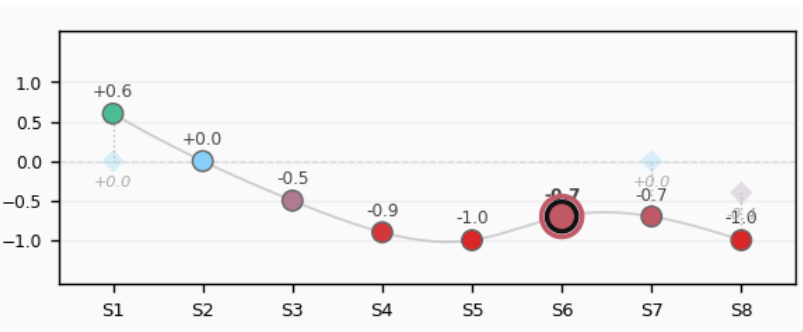}\\[2pt]
        \includegraphics[width=\textwidth]{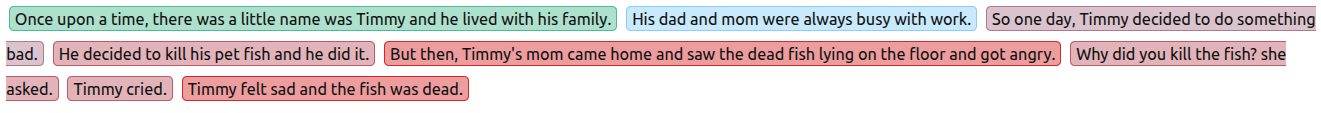}
        \caption{Shifting sentiment arc toward negative with 20 edit. Less-hued numbers show actual post-edit sentiment; larger gaps indicate lower faithfulness.}
        \label{app:fig:story_negative}
    \end{subfigure}\\[4pt]
    \begin{subfigure}{\textwidth}
        \centering
        \includegraphics[width=0.5\columnwidth]{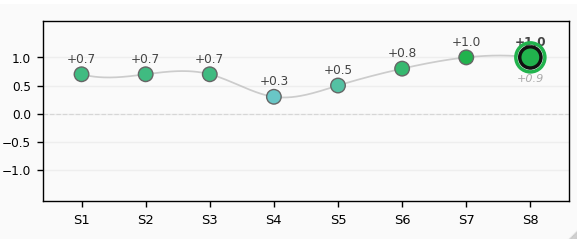}\\[2pt]
        \includegraphics[width=\textwidth]{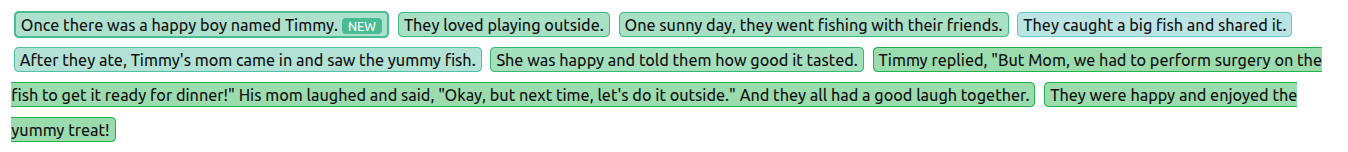}
        \caption{Shift Arc toward positive with 12 edits. Positive shifts required fewer edits reflecting the sentiment asymmetry in the quantitative results.}
        \label{app:fig:story_positive}
    \end{subfigure}
    \caption{Qualitative examples of iterative sentiment editing on a story (in-distribution). Each subfigure shows the sentiment arc (top) and annotated text (bottom).}
    \label{app:fig:story_qualitative}
\end{figure*}

Figure~\ref{app:fig:story_qualitative} illustrates the story editing process using our custom visualization tool. Figure~\ref{app:fig:story_original} shows the original sentiment trajectory, revealing a mix of positive, neutral, and negative tones that makes fine-grained control challenging. In Figure~\ref{app:fig:story_negative}, we shift the story toward a more negative tone without altering its core theme where a boy, a fish, and his mother contribute while transforming a harmless event into one involving a dead fish and an angry mother. This required 20 edits across all sentences. We observed stronger negative transformations sometimes necessitate modifying surrounding context, as locally extreme changes can conflict with the overall tone of prior or posterior sentences. Figure~\ref{app:fig:story_positive} shows the opposite transformation toward a more positive tone, requiring only 12 edits across 7 of 8 sentences, suggesting that positive sentiment shifts are more efficient when the original context already leans positive.

\begin{figure*}[!htbp]
    \centering
    \begin{subfigure}{\textwidth}
        \centering
        \includegraphics[width=0.5\columnwidth]{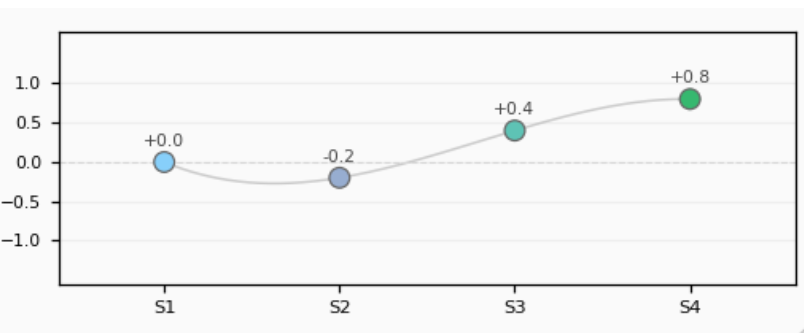}\\[2pt]
        \includegraphics[width=\textwidth]{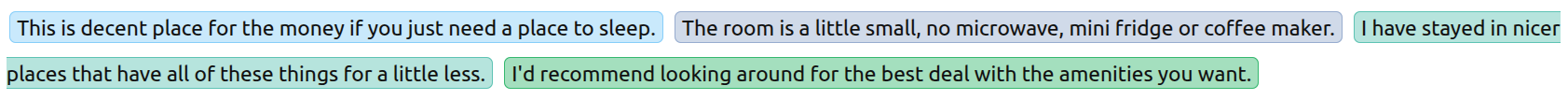}
        \caption{Original review and its sentiment arc.}
        \label{app:fig:review_original}
    \end{subfigure}\\[4pt]
    \begin{subfigure}{\textwidth}
        \centering
        \includegraphics[width=0.5\columnwidth]{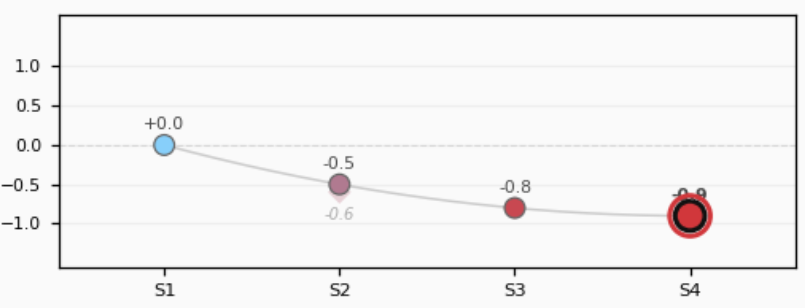}\\[2pt]
        \includegraphics[width=\textwidth]{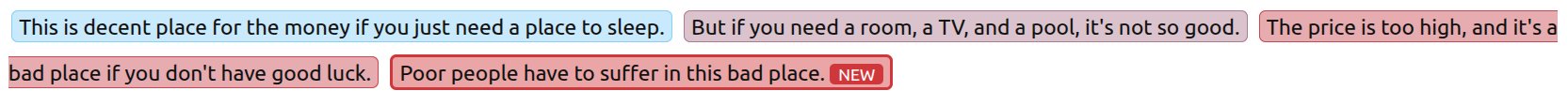}
        \caption{Shifting sentiment arc toward negative with 8 edits. Less-hued numbers show actual post-edit sentiment; larger gaps indicate lower faithfulness.}
        \label{app:fig:review_negative}
    \end{subfigure}\\[4pt]
    \begin{subfigure}{\textwidth}
        \centering
        \includegraphics[width=0.5\columnwidth]{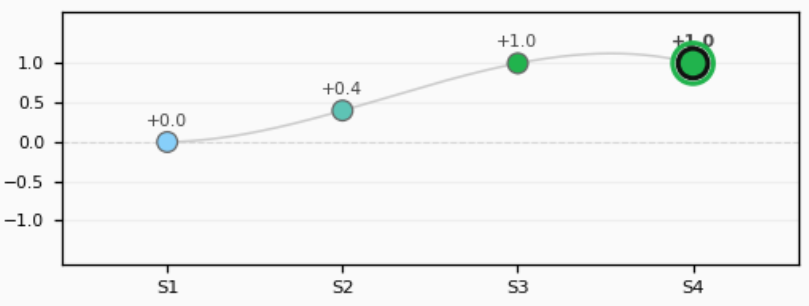}\\[2pt]
        \includegraphics[width=\textwidth]{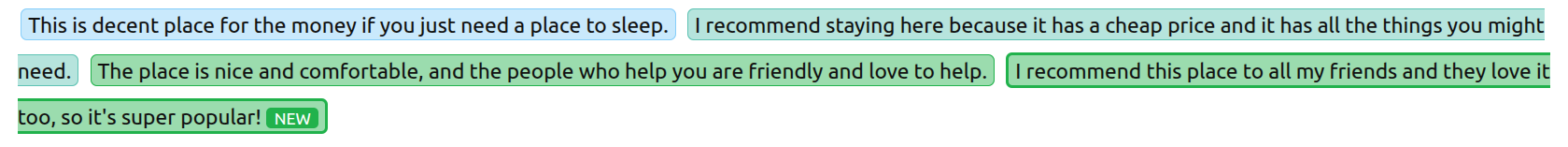}
        \caption{Shifting sentiment arc toward positive with 6 edits.}
        \label{app:fig:review_positive}
    \end{subfigure}
    \caption{Qualitative examples of iterative sentiment editing on a hotel review (out-of-distribution). Each subfigure shows the sentiment arc (top) and annotated text (bottom).}
    \label{app:fig:review_qualitative}
\end{figure*}

Figure~\ref{app:fig:review_qualitative} presents the same editing process applied to a hotel review as an out-of-distribution test. Figures~\ref{app:fig:review_qualitative} (a), (b), and (c) show the original review, negative arc shift (8 edits), and positive arc shift (6 edits) respectively. \mbl demonstrate strong OOD capability in review domain despite being trained exclusively on stories, maintaining coherent and fluent edits across both sentiment directions.

Overall, these examples demonstrate that \mbl achieves fine-grained controllable sentiment editing while preserving narrative coherence, with the number of required edits reflecting the asymmetry between negative and positive sentiment shifts observed in the quantitative results.

\subsection{Usage of AI}
\label{app:sec:ai_use}

We used Claude to assist with plotting and writing the HTML code for the user-study interface. We also used OpenAI models and Claude to revise the authors’ original writing for grammar and clarity.

% \section{Example Appendix}
% \label{sec:appendix}

% This is an appendix.

\end{document}